\documentclass[10pt,twocolumn,letterpaper]{article}

\usepackage{iccv}
\usepackage{times}
\usepackage{epsfig}
\usepackage{graphicx}
\usepackage{amsmath}
\usepackage{amssymb}
\usepackage{float}
\usepackage[caption = false]{subfig}
\usepackage{amsmath}
\usepackage{booktabs}

\DeclareMathOperator*{\argmin}{arg\,min}

\usepackage[pagebackref=true,breaklinks=true,letterpaper=true,colorlinks,bookmarks=false]{hyperref}

\newcommand*{\affaddr}[1]{#1} 
\newcommand*{\affmark}[1][*]{\textsuperscript{#1}}

 \iccvfinalcopy 

\ificcvfinal\fi
\begin{document}

\title{Generate, Segment and Refine: Towards Generic Manipulation Segmentation}

\author{Peng Zhou\affmark[1]\qquad Bor-Chun Chen\affmark[1]\qquad Xintong Han\affmark[2]\qquad Mahyar Najibi\affmark[1]\\ Abhinav Shrivastava\affmark[1]\qquad Ser Nam Lim\affmark[3]\qquad Larry S. Davis\affmark[1]\\
\affaddr{\affmark[1]University of Maryland, College Park \qquad}
\affaddr{\affmark[2]Malong Technologies \qquad}
\affaddr{\affmark[3]Facebook\\}
}

\maketitle

\begin{abstract}
  Detecting manipulated images has become a significant emerging challenge. The advent of image sharing platforms and the easy availability of advanced photo editing software have resulted in a large quantities of manipulated images being shared on the internet. While the intent behind such manipulations varies widely, concerns on the spread of fake news and misinformation is growing. Current state of the art methods for detecting these manipulated images suffers from the lack of training data due to the laborious labeling process. We address this problem in this paper, for which we introduce a manipulated image generation process that creates true positives using currently available datasets. Drawing from traditional work on image blending, we propose a novel generator for creating such examples. In addition, we also propose to further create examples that force the algorithm to focus on boundary artifacts during training. Strong experimental results validate our proposal.
\end{abstract}

\section{Introduction}

\begin{figure}[h!]
\centering
\includegraphics[width=1\linewidth]{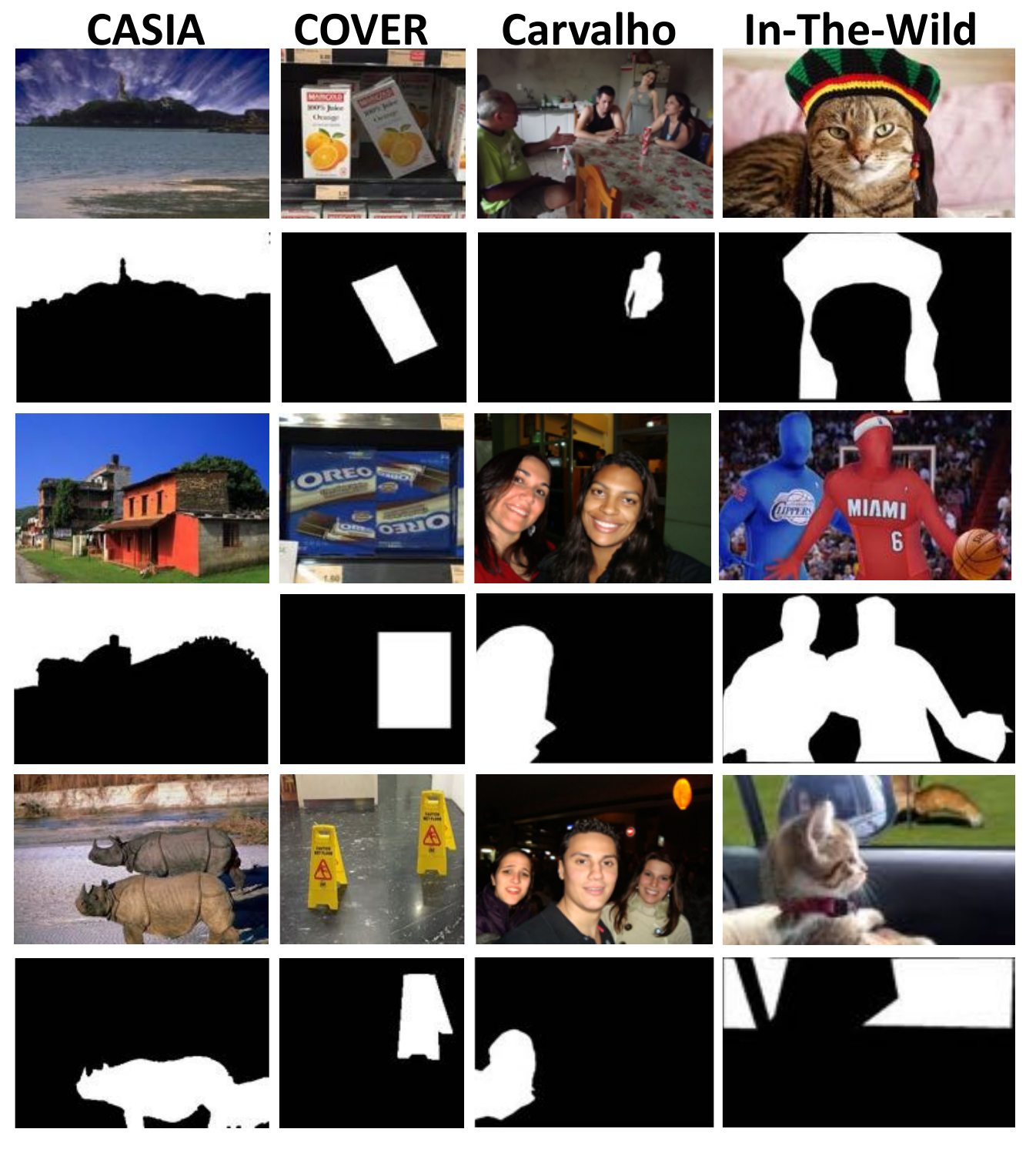}

   \caption{\textbf{Examples of manipulated images across different datasets.} Columns from left to right are images in CASIA~\cite{dong2013casia}, COVER~\cite{wen2016coverage}, Carvalho~\cite{de2013exposing}, and In-The-Wild~\cite{huh2018fighting}. The odd rows are manipulated images and the even rows are the ground truth masks. Different datasets contain different distributions (from animals to person), manipulation techniques (from copy-move (the second column) to splicing (the rest columns)) and post-processing methods (from no post-processing to various processes including filtering, illumination, and blurring).
   }
\label{fig:eg}
\end{figure}

\begin{figure*}[ht]
\begin{center}
   \includegraphics[width=0.96\linewidth]{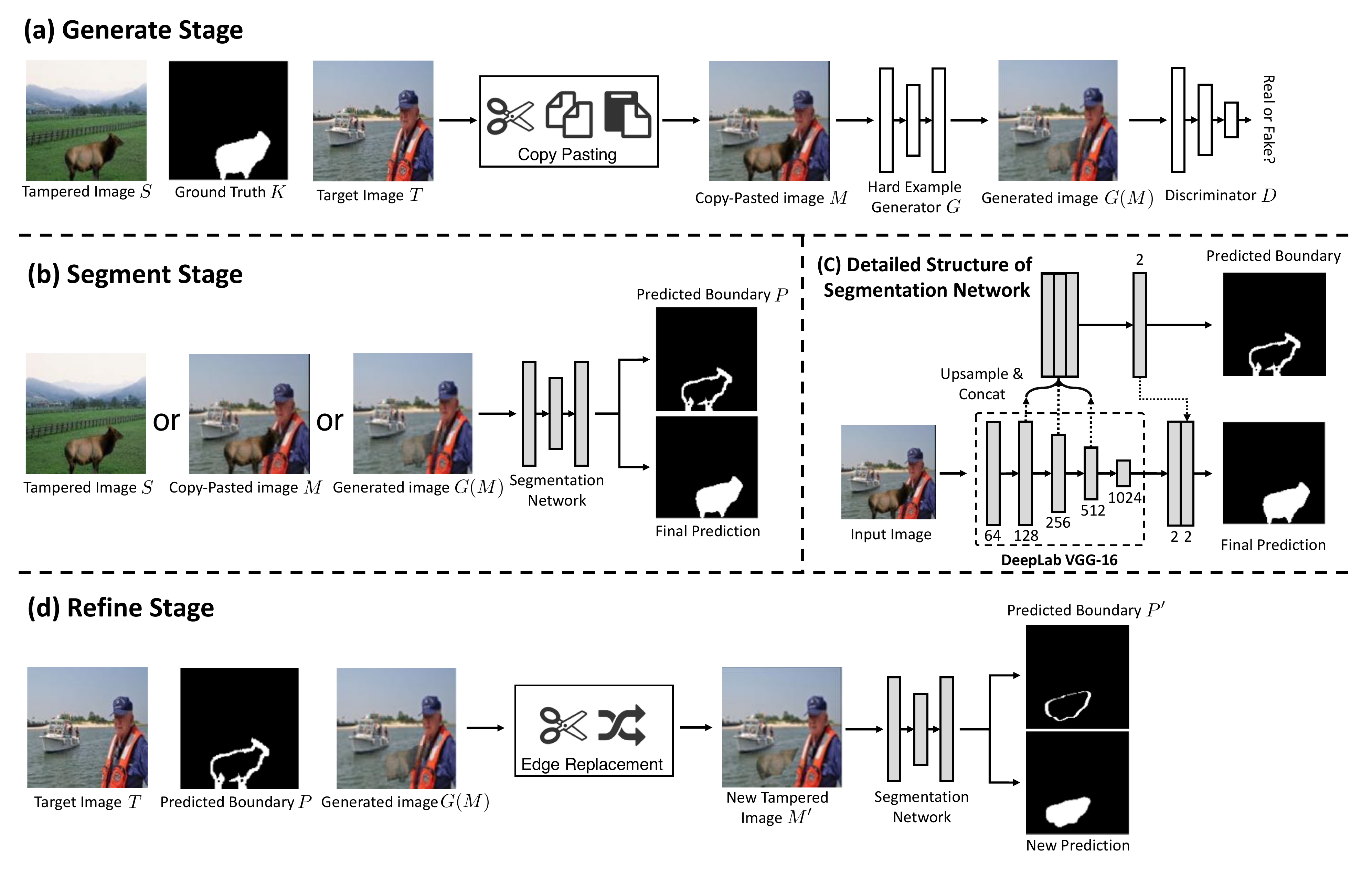}
\end{center}
   \caption{\textbf{GSR-Net framework overview}. \textbf{(a)} Given a tampered image $S$, an authentic target image $T$, and the ground truth mask $K$, the generation stage generates hard example $G(M)$ starting from a simple copy-pasting image $M$. \textbf{(b)} Feeding the training images, copy-pasted images or generated images as input, the segmentation stage learns to segment the boundary artifacts and fill the interior to produce the final prediction. \textbf{(c)} The segmentation network concatenates lower level features to predict boundary artifacts and then concatenate back the boundary feature to the segmentation branch for final prediction. \textbf{(d)} The refinement stage creates a novel tampered image with new boundary artifacts by replacing the predicted manipulated boundaries of segmentation stage with original authentic regions and learns to make a new prediction.}
\label{fig:frame}
\end{figure*}

Manipulated photos are becoming ubiquitous on social media due to the availability of advanced editing software, including powerful generative adversarial models such as \cite{isola2017image, yeh2017semantic}. While such images have been created for a variety of purposes, including memes, satires, etc., there are growing concerns on the abuse of manipulated images to spread fake news and misinformation. To this end, a variety of solutions have been developed towards detecting such manipulated images. 

While a number of proposed solutions posed the problem as a classification task \cite{cozzolino2018forensictransfer, zhou2017two}, where the goal is to classify whether a given image has been tampered with, there is great utility for solutions that are capable of detecting manipulated regions in a given image~\cite{huh2018fighting,zhou2017two,park2018double, salloum2018image}. In this paper, we similarly treat this problem as a semantic segmentation task and adapt GANs \cite{goodfellow2014generative} to generate samples to alleviate the lack of training data. The lack of training data has been an ongoing problem for training models to detect manipulated images. Scouring the internet for ``real'' tampered images~\cite{moreira2018image} is a laborious process that often also leads to over-fitting in the training process. Alternatively, one could employ a self-supervised process, where detected objects in one image are spliced onto another, with the caveat that such a process often results in training images that are not realistic. Of course, the best approach for generating training samples is to employ professional labelers to create realistic looking manipulated images, but this remains a very tedious process. It is therefore not surprising that existing datasets \cite{huh2018fighting, dong2010casia, dong2013casia, wen2016coverage, de2013exposing} are often not comprehensive enough to train models that generalize well.

Additionally, in contrast to standard semantic image segmentation, correctly segmenting manipulated regions depends more on visual artifacts that are often created at the boundaries of manipulated regions than on semantic content \cite{bappy2017exploiting, zhou2018learning}. Several challenges exist in recognizing these boundary artifacts. First, the space of manipulations is very diverse. One can, for example, do a {\em copy-move}, which copies and pastes image regions within the same image (the second column in Figure~\ref{fig:eg})
, or {\em splice}, which copies a region from one image and pastes it to another image (the remaining columns in Figure~\ref{fig:eg}). Second, a variety of post-processing such as compression, blurring, and various color transformations make it harder to detect boundary artifacts caused by tampering. See Figure~\ref{fig:eg} for some examples. Most existing methods \cite{huh2018fighting, zhou2018learning, park2018double, salloum2018image} that utilize discriminative features like image metadata, noise models, or color artifacts due to, for example, Color Filter Array (CFA) inconsistencies, have failed to generalize well for these reasons.

This paper introduces a two-pronged approach to (1) address the lack of comprehensive training data, as well as, (2) focus the training process on learning to recognize boundary artifacts better. We adopt GANs for addressing (1), but instead of relying on prior GAN methods~\cite{isola2017image, CycleGAN2017, karras2017progressive} that mainly explore image level manipulation, we introduce a novel objective function (Section~\ref{Generator}) that optimizes for the realism of the manipulated regions by blending tampered regions in existing datasets to assist segmentation. That is, given an annotated image from an existing dataset, our GAN takes the given annotated regions and optimizes via a blending based objective function to enhance the realism of the regions. Blending has been shown to be effective in creating training images effective for the task of object detection in \cite{debi2017cutpaste}, and this forms our main motivation in formulating our GAN.

To address (2), we propose a segmentation and refinement procedure. The segmentation stage localizes manipulated regions by learning to spot boundary artifacts. To further prevent the network from just focusing on semantic content, the refinement stage refines the predicted manipulation boundaries with authentic background and feed the new manipulated images back to the segmentation network. We will show empirically that the segmentation and refinement has the effect of focusing the model's attention on boundary artifacts during learning. (See Table~\ref{tab:ablation})

We design an architecture called GSR-Net which includes these three components---a generation stage, a segmentation stage and a refinement stage. The architecture of GSR-Net is shown in Figure~\ref{fig:frame}. During training, we alternatively train the generation GAN, followed by the segmentation and refinement stage, which take as input the output of the generation stage as well as images from the training datasets. The additional varieties of manipulation artifacts provided by both the generation and refinement stages produce models that exhibit very good generalization ability. We evaluate GSR-Net on four public benchmarks and show that it performs better or equivalent to state-of-the-art methods. Experiments with two different post-processing attacks further demonstrate the robustness of GSR-Net. In summary, the contributions of this paper are 1) A framework that augments existing datasets in a way that specifically addresses the main weaknesses of current approaches without requiring new annotations efforts; 2) Introducing a generation stage with a novel objective function based on blending for generating images effective for training models to detect tampered regions; 3) Introducing a novel refinement stage that encourages the learning of boundary artifacts inherent in manipulated regions, which, to the best of our knowledge, no prior work in this field has utilized to help training.

\section{Related Work}

\noindent\textbf{Image Manipulation Segmentation}. Park \etal.~\cite{park2018double} train a network to find JPEG compression discrepancies between manipulated and authentic regions. Zhou \etal.~\cite{zhou2017two, zhou2018learning} harness noise features to find inconsistencies within a manipulated image. Huh and Liu \etal.~\cite{huh2018fighting} treat the problem as anomaly segmentation and use metadata to locate abnormal patches. The features used in these works are based on the assumption that manipulated regions are from a different image, which is not the case in copy-move manipulation. However, our method directly focuses on general artifacts in the RGB channel without specific feature extraction and thus can be applied to copy-move segmentation. More related works of Salloum \etal.~\cite{salloum2018image} and Bappy \etal.  \cite{bappy2017exploiting} show the potential of boundary artifacts in different image manipulation techniques. Salloum \etal.~\cite{salloum2018image} adopt a Multi-task Fully Convolutional Network (MFCN) \cite{long2015fully} to manipulation segmentation by providing both segmentation and edge annotations. Bappy \etal.~\cite{bappy2017exploiting} design a Long Short-Term Memory (LSTM) \cite{hochreiter1997long} based network to identify RGB boundary artifacts at both the patch and pixel level. These methods are sources of motivation for us to exploit boundary artifacts as a strong cue for detecting manipulations. 

\noindent\textbf{GAN Based Image Editing}.
GAN based image editing approaches have witnessed a rapid emergence and impressive results have been demonstrated recently \cite{tsai2017deep,lalonde2007using,wang2017high, karras2017progressive, CycleGAN2017}. Prior and concurrent works force the output of GAN to be conditioned on input images through extra regression losses (\eg., $\ell_2$ loss) or discrete labels. However, these methods manipulate the whole images and do not fully explore region based manipulation. In contrast, our GAN manipulates minor regions and help better generalization ability of a manipulation segmentation network. A more related work is Tsai \etal.~\cite{tsai2017deep} that generates natural composite images using both scene parsing and harmonized ground truth. Even though it targets on region manipulation, Section \ref{ablate} shows that our method performs better in terms of assisting segmentation.

\noindent\textbf{Adversarial Training}. Discriminative feature learning has motivated recent research on adversarial training on several tasks. Shrivastava \etal.~\cite{shrivastava2017learning} propose a simulated and unsupervised learning approach which utilizes synthetic images to generate realistic images. Wang \etal.~\cite{wang2017fast} boost the performance on occluded and deformed objects through an online hard negative generation network. Wei \etal.~\cite{wei2017object} investigate an adversarial erasing approach to learn dense and complete semantic segmentation. Le \etal.~\cite{le2018a+} propose an adversarial shadow attenuation network to make correct predictions on hard shadow examples. However, their approach is difficult to adapt to manipulation segmentation because they either generate whole synthetic images or introduce new manipulation such as erasing. In contrast, we refine manipulated regions with original ones to assist segmentation network.

\section{Approach}

We describe the GSR-net in details in the following sections. Key to the generation is the utilization of a GAN with a loss function central around using blending to optimize for producing realistic training images. The segmentation and refinement stage are specially designed to single out boundaries of the manipulated regions in order to guide the training process to pay extra attention to boundary artifacts. 

\subsection{Generation}\label{Generator}



\noindent\textbf{Generator.} Referring to Figure~\ref{fig:frame} (a), the generator is given as input both copy-pasted images and ground truth masks. To prepare the input images, we start with the training samples in manipulation datasets (\eg., CASIA 2.0 \cite{dong2013casia}). Given a training image $S$, the corresponding ground truth binary mask $K$ and an authentic target image $T$ from a clean dataset (\eg., COCO \cite{lin2014microsoft}), we first create a simple copy-pasted image $M$ by taking $S$ as foreground and $T$ as background:

\begin{equation}
    M = K \odot S + (1-K) \odot T,
\end{equation}
where $\odot$ represents pointwise multiplication.

In Poisson blending \cite{perez2003poisson}, the final value of pixel $i$ in the manipulated regions is 
\begin{align}
    b_i & = \displaystyle \argmin_{b_i} \sum_{s_i\in S,  \mathcal{N}_i \subset S }{||\nabla b_i-\nabla s_i||_2} \nonumber \\ 
    & + \sum_{s_i\in S, \mathcal{N}_i \not \subset S}{||b_i-t_i||_2}, 
\end{align}
where $\nabla$ denotes the gradient, $\mathcal{N}_i$ is the neighborhood (\eg., up, down, left and right) of the pixel at position $i$, $b_i$ is the pixel in the blended image $B$, $s_i$ is the pixel in $S$ and $t_i$ is the pixel in $T$. 

Similar to Poisson blending, we optimize the generator to blend neighborhoods in the resulting image that now contains copy-pasted regions and background regions. A key part of our loss function enforces the shapes of the tampered regions, while maintaining the background regions. To maintain background regions, we utilize $\ell_1$ loss as in \cite{isola2017image} to reconstruct the background:
\begin{equation}
L_\text{bg}=\frac{1}{N_\text{bg}}\sum_{t_i\in T, k_i=0}||m_i-t_i||_1, 
\end{equation}
where $N_\text{bg}$ is the total number of pixels in the background, $m_i$ is the pixel in $M$ and $k_i$ is the value in mask $K$ at position $i$. To maintain the shape of manipulated regions, we apply a Laplacian operator to the pasted regions and reconstruct the gradient of this region to match the source region:
\begin{equation}
L_\text{grad}=\frac{1}{N_\text{fg}}\sum_{s_i\in S, k_i=1}||\Delta m_i-\Delta s_i||_1, 
\end{equation}
where $\Delta$ denotes the Laplacian operator and $N_\text{fg}$ is the total number of pixels in pasted regions. To further constrain the shape of pasted regions, we add an additional edge loss as denoted by
\begin{equation}
L_\text{edge}=\frac{1}{N_\text{edge}}\sum_{s_i\in S, e_i=1}||m_i-s_i||_1, 
\end{equation}
where $N_\text{edge}$ is the number of boundary pixels and  $e_i$ is the value of the edge mask at position $i$, which is obtained by the absolute difference between a dilation and an erosion on $K$. To generate realistic manipulated images, we add an adversarial loss $L_\text{adv}$, as explained below, that serves to encourage the generator to produce increasingly realistic images as the training progresses.

\smallskip
\noindent\textbf{Discriminator.} In our discriminator, a crucial detail to point out is that the manipulated regions are typically occupying only a small area in the image. Hence, it is beneficial to restrict the GAN discriminator's attention to the structure in local images patches. This is reminiscent of ``PatchGAN''~\cite{isola2017image} that only penalizes structure at the scale of patches. Similar to PatchGAN, our discriminator applies a final fully convolutional layer at a patch scale of $N \times N$. The discriminator distinguishes the authentic image $T$ as real and the generated image $G(K,M)$ as fake by maximizing:
\begin{align}
L_\text{adv} & =\mathbb{E}_{T}{[\log(D(K,T))]} \nonumber
 \\
 & + \mathbb{E}_{M}{[1-\log(D(K,G(K,M)))]}, 
\end{align}
where $K$ is concatenated with $G(K,M)$ or $T$ as the input to the discriminator.

\smallskip
 The final loss function of the generator is given as
\begin{equation}
L_G =L_\text{bg} + \lambda_\text{grad} L_\text{grad} + \lambda_\text{edge}  L_\text{edge} + \lambda_\text{adv} L_\text{adv}, 
\end{equation}
where $\lambda_\text{grad}$, $\lambda_\text{edge}$, and $\lambda_\text{adv}$ are parameters which control the importance of the corresponding loss terms.
Conditioned on this constraint, the generator preserves background and texture information of pasted regions while blending the manipulated regions with the background.

\subsection{Segmentation}

For segmentation, we simply adopt the publicly available VGG-16 \cite{simonyan2014very} based DeepLab model \cite{chen2018deeplab}. The network structure is depicted in Figure~\ref{fig:frame} (c), consisting of a boundary branch predicting the manipulated boundaries and a segmentation branch predicting the interior. In particular, to enhance attention on boundary artifacts, we introduce boundary information by subtracting the erosion from the dilation of the binary ground truth mask to obtain the boundary mask. We then predict this boundary mask through concatenating bi-linearly up-sampled intermediate features and passing them to a $1 \times 1$ convolutional layer to form the boundary branch. Finally, we concatenate the output features of the boundary branch with the up-sampled features of the segmentation branch. Empirically, we noticed such multi-task learning helps the generalization of the final model. Only the segmentation branch output is used for evaluation during inference. During training, we select the copy-pasted examples $M$, generated examples $G(M)$ and training samples $S$ in the dataset as input to the segmentation network which provides a larger variety of manipulation. The loss function of the segmentation network is an average, two class softmax cross entropy loss.

\subsection{Refinement}

The goal of the refinement stage is to draw attention to the boundary artifacts during training, taking into account the fact that boundary artifacts play a more pivotal role than semantic content in detecting manipulations \cite{bappy2017exploiting, zhou2018learning}. While we may be able to employ prior erasing based adversarial mining methods \cite{wei2017object, wang2017fast}, they are not suitable for our purpose because it will introduce new erasing artifacts. Instead, the refinement stage utilizes the prediction of the segmentation stage to produce new boundary artifacts through replacing with original regions. As illustrated in Figure~\ref{fig:frame} (d), given an authentic target image $T$ in which the manipulated regions was inserted, the manipulated image $M$ (which could also be the generated image $G(M)$), and the manipulated boundary prediction $P$ by the segmentation stage, we replace the pixels in predicted boundaries by the authentic regions in $T$ and create a novel manipulated image:

\begin{equation}
    M' = T \odot P + M \odot (1-P), 
\end{equation}
where $M'$ is the novel manipulated image with new boundary artifacts. The corresponding segmentation ground truth now becomes 
\begin{equation}
    K' = K - K \odot P, 
\end{equation} where $K'$ is the new manipulated mask for $M'$. The new boundary artifact mask can be extracted in the same way as the previous step. Notice that the refinement stage utilizes the target images $T$ to help training, providing more side information to spot the artifacts. Taking as input the new manipulated images, the same segmentation network then learns to predict the new manipulated boundaries and interior regions. 

Similar to \cite{wei2017object}, multiple refinement operations are possible and there is a tradeoff between training time and performance. However, the difference is that the segmentation network in the refinement stage shares weights with that in the segmentation stage. The weight sharing enables us to use a single segmentation network at inference. As a result, the network learns to focus more attention on boundary artifacts with no additional cost at inference time. 

\section{Experiments}
We evaluate the performance of GSR-Net on four public benchmarks and compare it with the state-of-the-art methods. We also analyze its robustness under several forms of post-processing.

\subsection{Implementation Details}

GSR-Net is implemented in Tensorflow \cite{abadi2016tensorflow} and trained on an NVIDIA GeForce TITAN P6000. The input to the generation network (both generator and discriminator) is resized to $256 \times 256$. The generator is based on U-net \cite{ronneberger2015u} and the discriminator has the same structure as $70 \times 70$ PatchGAN \cite{isola2017image} without the batch normalization layers . We add batch normalization \cite{ioffe2015batch} to the DeepLab VGG-16 model. The segmentation network is fine-tuned from ImageNet \cite{deng2009imagenet} pre-trained weights and the generation network is trained from scratch. We use Adam \cite{kingma2014adam} optimizer with a fixed learning rate of $1 \times 10^{-4}$ for all the subnetworks. Optimizer of the generator, discriminator and segmentation network are updated in an alternating fashion. To avoid overfitting, weight decay with a factor of $5 \times 10^{-5}$ and $50\%$ dropout \cite{srivastava2014dropout} are applied on the segmentation network. The hyperparameters of $(\lambda_\text{grad},\lambda_\text{edge},\lambda_\text{adv})$ are set to $(1,2,5)$ empirically to balance the loss values. Only random flipping augmentation is applied during training. We feed the copy-pasted images, generated images, and the training samples to the segmentation network. We train the whole network jointly for 50K iterations with a batch size of 4.

Only the segmentation network is used for inference and we use the segmentation branch as the final prediction. Due to the small batch size used during training, we utilize instance normalization as in \cite{isola2017image} on every test image. After prediction, instead of using mean-shift as in \cite{huh2018fighting}, we simply dilate and threshold connected components to remove small noisy particles.

\begin{table*}[t]
\centering
\begin{tabular}{p{6cm}ccccccccc}
 \multicolumn{1}{l}{Dataset}
  & \multicolumn{2}{c}{\textbf{Carvalho}}  & \multicolumn{2}{c}{\textbf{In-The-Wild}}  & \multicolumn{2}{c}{\textbf{COVER}}     & \multicolumn{2}{c}{\textbf{CASIA}} \\ 
  \hline
 Metrics & {MCC} & {F1}& {MCC} & {F1}& {MCC} & {F1}& {MCC} & {F1}\\ 
 \hline
NOI~\cite{mahdian2009using}  &0.255 &0.343 &0.159 &0.278 & 0.172 &0.269 & 0.180 &0.263     \\
CFA~\cite{ferrara2012image}   &0.164 & 0.292 &0.144 &0.270 & 0.050 &0.190 & 0.108 &0.207       \\
MFCN~\cite{salloum2018image} &0.408 & 0.480 &-  &- & - & - &0.520 & 0.541   \\ 
RGB-N~\cite{zhou2018learning} &0.261 & 0.383     &0.290   &0.424    & 0.334 &0.379 &0.364                     & 0.408        \\
EXIF-consistency~\cite{huh2018fighting}*   &0.420 &0.520 &0.415  & 0.504 & 0.102 & 0.276 &0.127 &0.204\\
DeepLab (baseline) &0.343 &0.420 &0.352 & 0.472 &0.304 &0.376 &0.435 & 0.474 \\
\textbf{GSR-Net (ours)}& \textbf{0.462}   & \textbf{0.525} &\textbf{0.446}  &\textbf{0.555} &\textbf{0.439}   &    \textbf{0.489}         &      \textbf{0.553}     & \textbf{0.574}   \\
\end{tabular}

\caption{\textbf{$MCC$ and $F_1$ score comparison on four standard datasets.} `-' denotes that the result is not available in the literature. * Our method is $1600$ times faster than EXIF-consistency.}
\label{tab:f1}
\end{table*}

\subsection{Datasets and Experiment Setting}

\noindent\textbf{Datasets}. We evaluate our performance on the following four datasets:

\noindent$\bullet$\enskip\textbf{In-The-Wild \cite{huh2018fighting}}: In-The-Wild is a splicing dataset with 201 spliced images collected online. The annotation is manually created by \cite{huh2018fighting} and the manipulated regions are usually person and animal. 

\noindent$\bullet$\enskip\textbf{COVER \cite{wen2016coverage}}: COVER focuses on copy-move manipulation and has 100 images. The manipulation objects are used to cover similar objects in the original authentic images and thus are challenging for humans to recognize visually without close inspection. 

\noindent$\bullet$\enskip\textbf{CASIA \cite{dong2013casia,dong2010casia}}: CASIA has two versions. CASIA 1.0 contains 921 manipulated images including splicing and copy move. The objects are carefully selected to match with the context in the background. Cropped regions are subjected to post-processing including rotation, distortion, and scaling. CASIA 2.0 is a more complicated dataset with 5123 images. Manipulations include splicing and copy move. Post-processing like filtering and blurring is applied to make the regions visually realistic. The manipulated regions cover animals, textures, natural scenes, \etc. We use CASIA 2.0 to train our network and test it on CASIA 1.0 in Section \ref{main}. 

\noindent$\bullet$\enskip\textbf{Carvalho \cite{de2013exposing}}: Carvalho is a manipulation dataset designed to conceal illumination differences between manipulated regions and authentic regions. The dataset contains 100 images and all the manipulated objects are people. Contrast and illumination are adjusted in a post-processing step.

\noindent\textbf{Evaluation Metrics}. We use pixel-level F1 score and MCC as the evaluation metrics when comparing to other approaches. For fair comparison, following the same measurement as \cite{salloum2018image, huh2018fighting, zhou2018learning}, we vary the prediction threshold to get binary prediction mask and report the optimal score over the whole dataset.

\begin{table}[t]
\centering  
\small
\begin{tabular}{p{1.8cm}ccccc} 
 Dataset & Carvalho & In-the-Wild & COVER & CASIA\\   
 \hline
DeepLab &0.420 &0.472  & 0.376 &0.474\\
DL + CP &0.446 &0.504  &0.410 &0.503 \\
DL + G  &0.460 &0.524 &0.434 &0.506 \\
DL + DIH &0.384 &0.421  &0.342 &0.420 \\
DL + CP + G  &0.472 &0.528 & 0.444 & 0.507     \\
GS-Net   &0.515 &0.540  &0.455  &0.545 \\
GSR-Net &0.525 &0.555 &0.489 &0.574\\
\end{tabular}

\caption{\textbf{Ablation analysis on four datasets.} Each entry is the F1 score tested on individual dataset.}
\label{tab:ablation}
\end{table}

\subsection{Main Results} \label{main}

In this section, We present our results for the task of manipulation segmentation. We fine-tune our model on CASIA 2.0 from the ImageNet pre-trained model and test the performance on the aforementioned four datasets. We compare with methods described below:

\noindent$\bullet$\enskip\textbf{NoI}  \cite{mahdian2009using}:
A noise inconsistency method which predicts regions as manipulated where the local noise is inconsistent with authentic regions. We use the code provided by Zampoglou \etal.~\cite{zampoglou2017large} for evaluation.

\noindent$\bullet$\enskip\textbf{CFA} \cite{ferrara2012image}:
A CFA based method which estimates the internal CFA pattern of the camera for every patch in the image and segments out the regions with anomalous CFA features as manipulated regions. The evaluation code is based on Zampoglou \etal.~\cite{zampoglou2017large}.

\noindent$\bullet$\enskip\textbf{RGB-N} \cite{zhou2018learning}:
A two-stream Faster R-CNN based approach which combines features from the RGB and noise channel to make the final prediction. We train the model on CASIA 2.0 using the code provided by the authors \footnote{\url{https://github.com/pengzhou1108/RGB-N}}. 

\noindent$\bullet$\enskip\textbf{MFCN}     \cite{salloum2018image}:
A multi-task FCN based method which harnesses both an edge mask and segmentation mask for manipulation segmentation. Hole filling is applied for the edge branch to make the prediction. The final decision is the intersection of the two branches. We directly report the results from the paper since the code is not publicly available.

\noindent$\bullet$\enskip\textbf{EXIF-consistency}   \cite{huh2018fighting}:
A self-consistency approach which utilizes metadata to learn features useful for manipulation localization. The prediction is made patch by patch and post-processing like mean-shift \cite{cheng1995mean} is used to obtain the pixel-level manipulation prediction. We use the code provided by the authors \footnote{\url{https://github.com/minyoungg/selfconsistency}} for evaluation.

\noindent$\bullet$\enskip\textbf{DeepLab}:
Our baseline model which directly adopts DeepLab VGG-16 model to manipulation segmentation task. No generation, boundary branch or refinement stage is added.

\noindent$\bullet$\enskip\textbf{GSR-Net}:
Our full model combining generation, segmentation and refinement for manipulation segmentation.

The final results, presented in Table~\ref{tab:f1}, highlight the advantage of GSR-Net. For supervised methods \cite{zhou2018learning,salloum2018image}, we train the model on CASIA 2.0 and evaluate on all the four datasets. For other unsupervised methods \cite{mahdian2009using,ferrara2012image, huh2018fighting}, we directly test the model on all datasets. GSR-Net outperforms other approaches by a large margin on COVER, suggesting the advantage of our network on copy-move manipulation. Also, GSR-Net has an improvement on CASIA 1.0 and Carvalho. Moreover, in terms of computation time, EXIF-consistency takes $160$ times more computation (80 seconds for an $800\times 1200$ image on average) than ours (0.5s per image). Compared to boundary artifact based methods, our GSR-Net outperforms MFCN by a large margin, indicating the effectiveness of the generation and refinement stages. In addition to that, no hole filling is required since our approach does not perform late fusion with the boundary branch, but utilizing boundary artifacts to guide the segmentation branch instead.

Our method outperforms the baseline model by a large margin, showing the effectiveness of the proposed generation, segmentation and refinement stages.

\begin{figure}[t!]
\begin{center}
  \includegraphics[width=\linewidth]{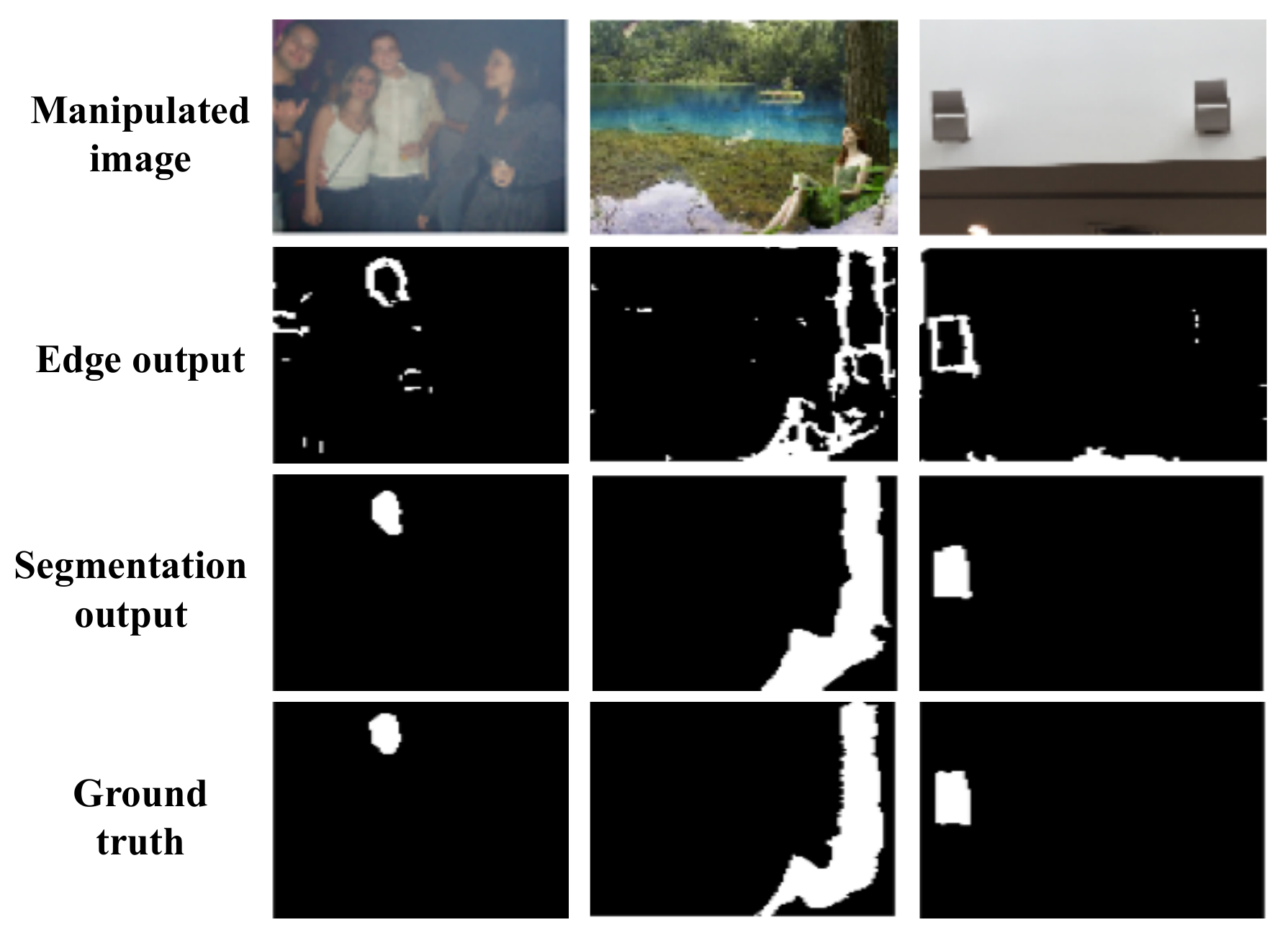}
\end{center}
   \caption{Qualitative visualization. The first row shows manipulated images on different datasets. The second indicates the final manipulation segmentation prediction. The third row illustrates the output of boundary artifacts branch. The last row is the ground truth. }
\label{fig:qr}
\end{figure}

\subsection{Ablation Analysis}\label{ablate}

We quantitatively analyze the influence of each component in GSR-Net in terms of F1 score.

\noindent$\bullet$\enskip\textbf{DL + CP}:
DeepLab VGG-16 model with just the segmentation output, using simple copy-pasted (no generator) and CASIA 2.0 images  during training.

\noindent$\bullet$\enskip\textbf{DL + G}:
DeepLab VGG-16 model with just the segmentation output, using generated and CASIA 2.0 images during training.

\noindent$\bullet$\enskip\textbf{DL + \cite{tsai2017deep}}:
 DeepLab VGG-16 model with just the segmentation output, using the images generated from \cite{tsai2017deep} and CASIA 2.0 images during training. We adapt deep harmonization network for the generation stage as it also manipulate regions.

\noindent$\bullet$\enskip\textbf{DL + CP + G}:
DeepLab VGG-16 model with just the segmentation output, using both copy-pasted, generated and CASIA 2.0 images during training.

\noindent$\bullet$\enskip\textbf{GS-Net}:
Generation and segmentation network with boundary artifact guided manipulation segmentation. No refinement stage is incorporated.

\begin{figure}[t!]
\begin{center}
   \includegraphics[width=1\linewidth]{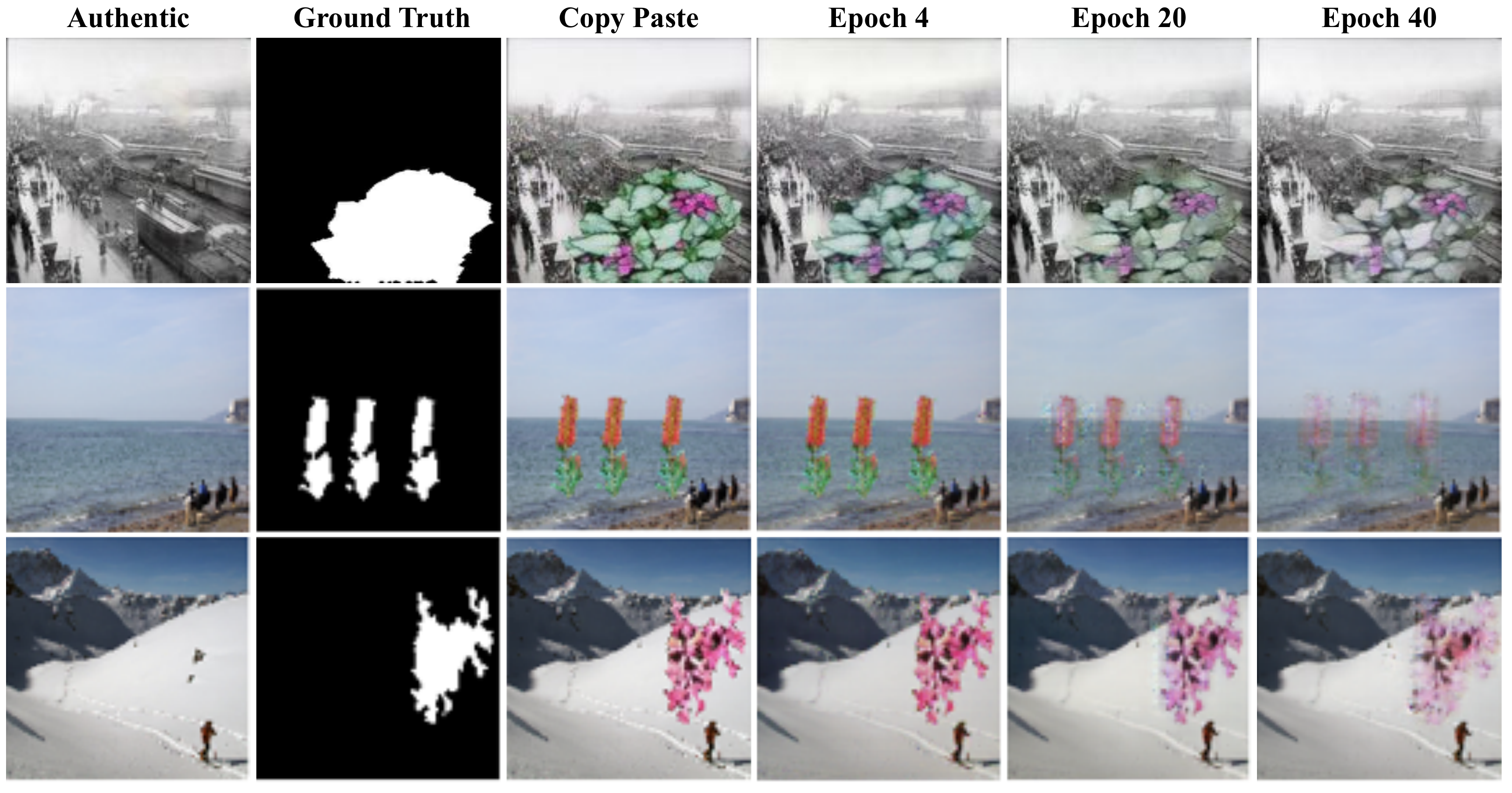}
\end{center}
   \caption{Qualitative visualization of the generation network. The first two columns show the authentic background and manipulation mask. As the number of epochs increases, the manipulated region matches better with the background and thus boundary artifacts are harder to identify.}
\label{fig:gan}
\end{figure}

 \begin{figure*}[t!]
\begin{center}
   \subfloat[In-The-Wild JPEG attack]{\includegraphics[width=0.25\linewidth, height=0.2\linewidth]{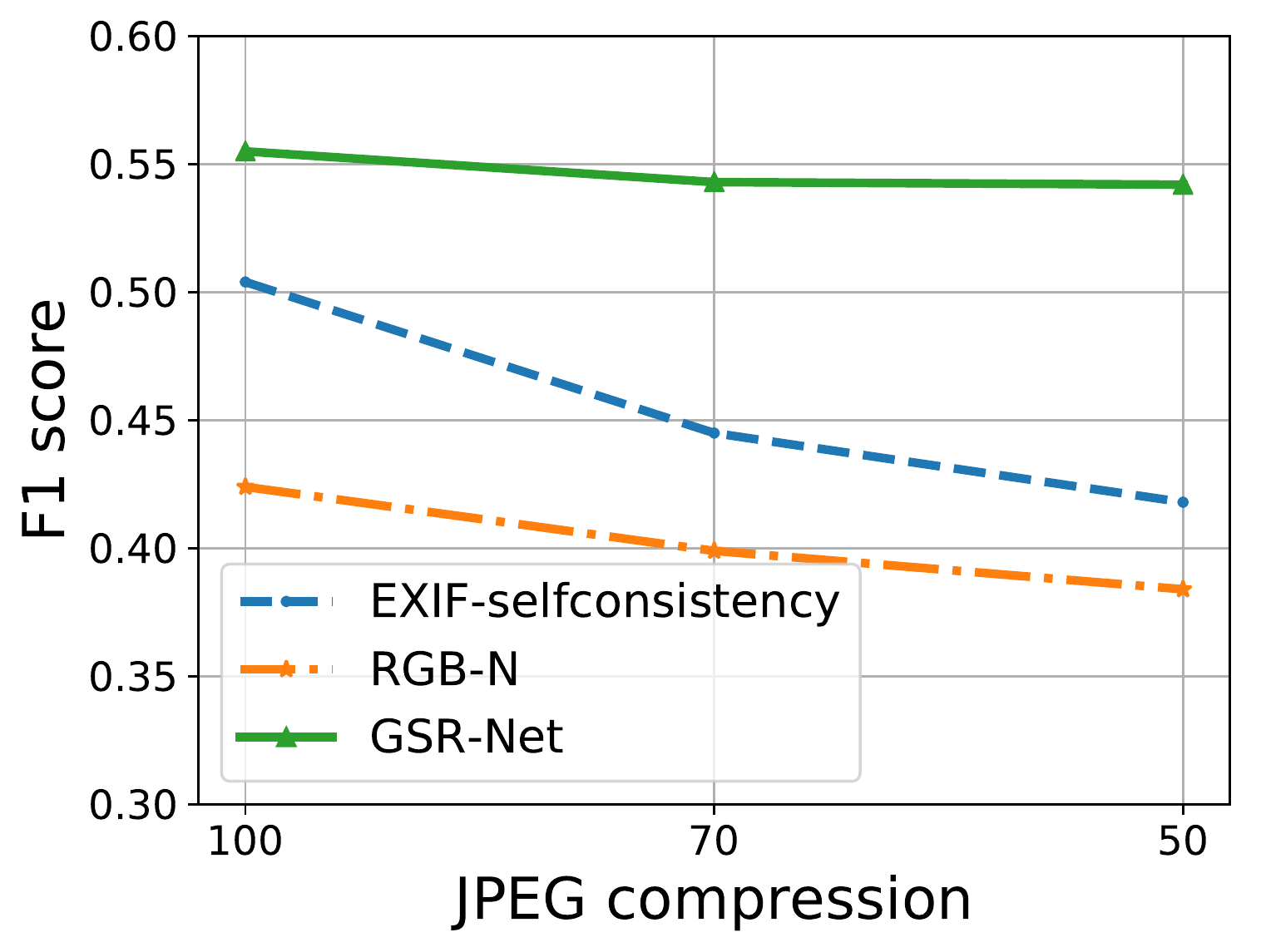}}
   \subfloat[In-The-Wild scale attack]{\includegraphics[width=0.25\linewidth, height=0.2\linewidth]{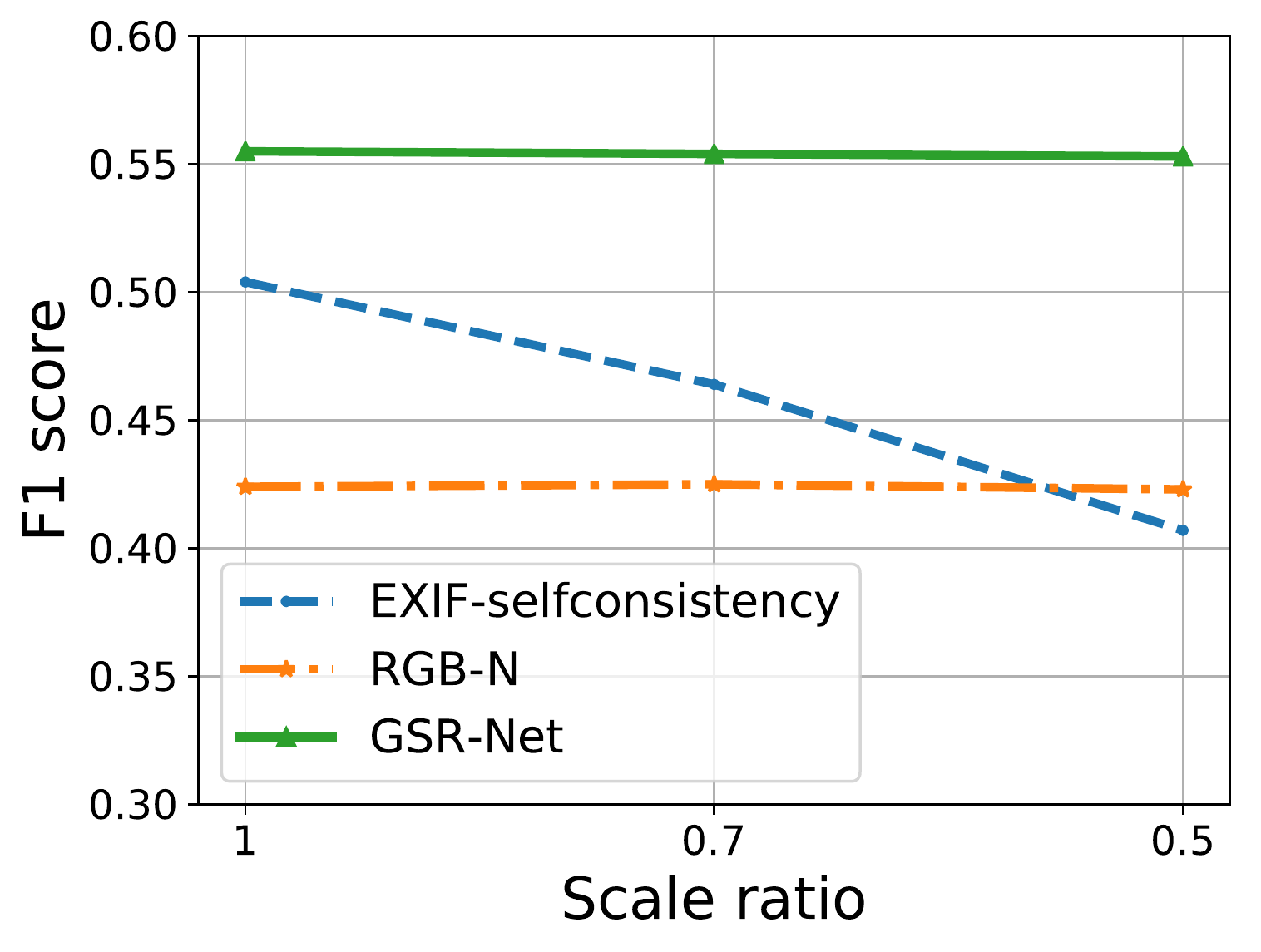}}
   \subfloat[Carvalho JPEG attack]{\includegraphics[width=0.25\linewidth, height=0.2\linewidth]{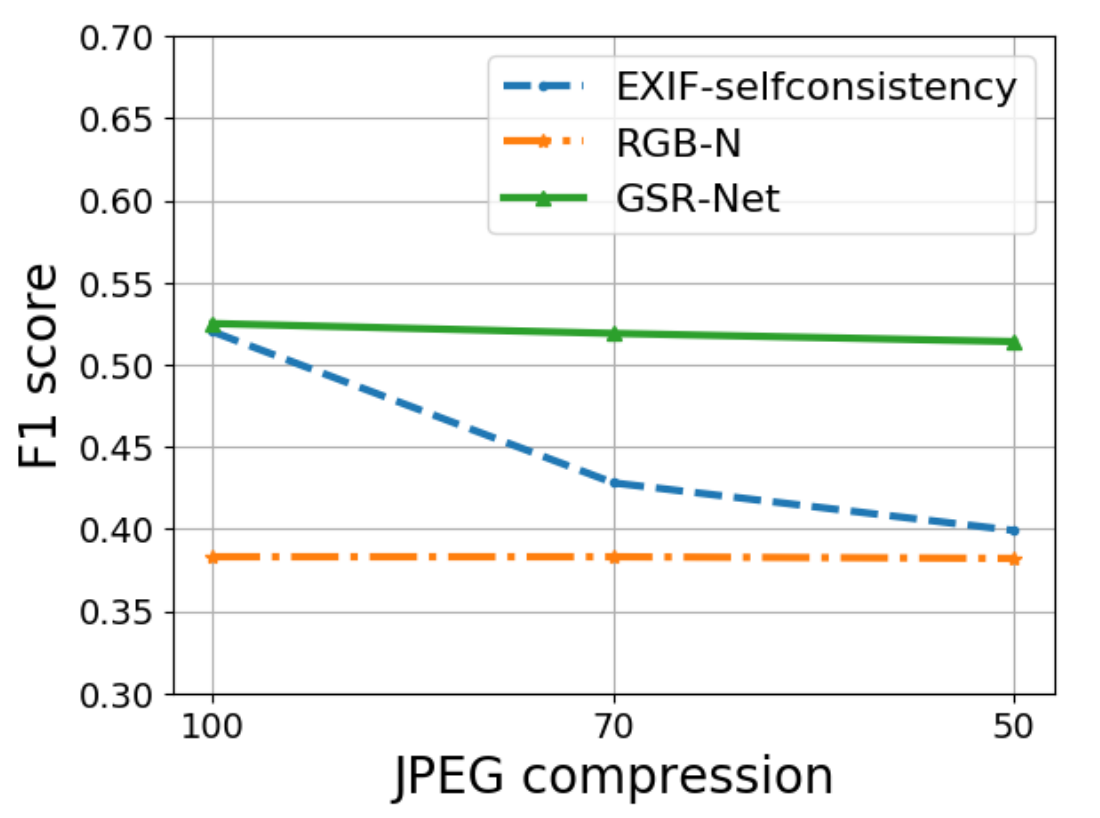}}
   \subfloat[Carvalho scale attack]{\includegraphics[width=0.25\linewidth,height=0.2\linewidth]{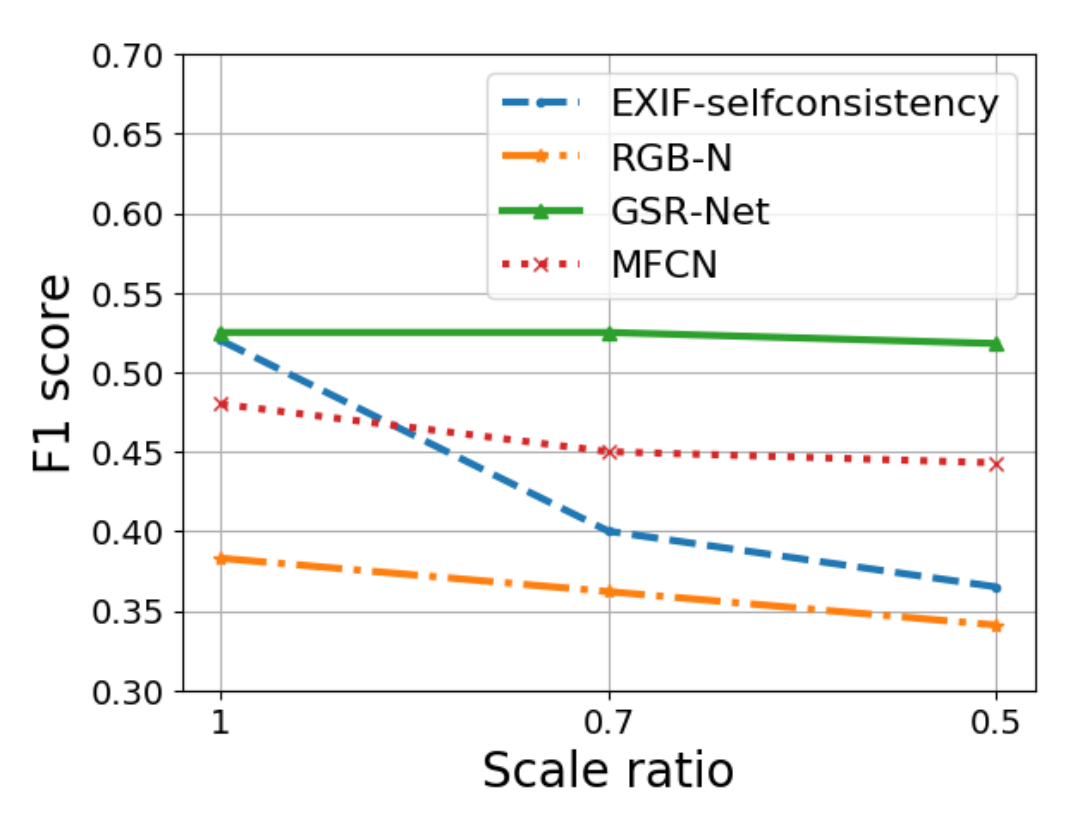}}
\end{center}
   \caption{Analysis of robustness under different attacks. Attacks with JPEG compression consists of quality factors of 70 and 50; scale attacks use scaling ratios of 0.7 and 0.5. (a) JPEG compression attacks on In-The-Wild. (b) Scale attacks on In-The-Wild. (c) JPEG compression attacks on Carvalho. (d) Scale attacks on Carvalho.}
\label{fig:attack}
\end{figure*}

The results are shown in Table~\ref{tab:ablation}. Starting from our baseline model, simply adding copy-pasted images (\textbf{DL + CP}) achieves improvement due to broadening the manipulation distribution. In addition, replacing copy-pasted images with generated images (\textbf{DL + G}) also shows improvement compared to \textbf{DL + CP} on most of the datasets as it refines the boundary from naive copy-pasting.  As expected, adding both copy-pasted images and generated hard examples (\textbf{DL + CP + G}) is more useful because the network has access to a larger distribution of manipulation.

Compared to applying deep harmonization network \textbf{DL + \cite{tsai2017deep}}), our generation approach (\textbf{DL + G}) performs better as it aligns well with the natural process of manipulation.

The results also indicate the impact of boundary guided segmentation network. 
Directly predicting segmentation (\textbf{DL + CP + G}) does not explicitly learn manipulation artifacts, and thus has limited generalization ability compared to \textbf{GS-Net}.  Furthermore, \textbf{GSR-Net} boosts the performance on \textbf{GS-Net} since the refinement stage introduces new boundary artifacts. 

\subsection{Robustness to Attacks} \label{robust}

We apply both JPEG compression and image scaling attacks to test images of  In-The-Wild and Carvalho datasets. We compare GSR-Net with RGB-N \cite{zhou2018learning} and EXIF-selfconsistency \cite{huh2018fighting} using their publicly available code, and MFCN \cite{salloum2018image} using the numbers reported in their paper. Figure~\ref{fig:attack} shows the results, which indicates our approach yields more stable performance than prior methods.

\subsection{Segmentation with COCO Annotations}

This experiment shows how much gain our model achieves without using the manipulated images in CASIA 2.0. Instead of carefully manipulated training data, we only utilize the object annotations in COCO to create manipulated images. We compare the result of using different training data as follows:

\noindent$\bullet$\enskip\textbf{CP + S}:
Only using copy-pasted images to train the segmentation network.

\noindent$\bullet$\enskip\textbf{CP + G + S}:
Using both copy-pasted and generated images.

\noindent$\bullet$\enskip\textbf{CP + G + SR}:
Using copy-pasted images and generated images. The refinement stage is applied.

Results are presented in Table~\ref{tab:weakly}. The performance using only copy-pasted images (\textbf{CP + S}) on the four datasets indicates that our network truly learns boundary artifacts. Also, the improvement after adding generated images (\textbf{CP + G + S}) shows that our generation network provides useful manipulation examples that increases generalization. Last, the refinement stage (\textbf{CP + G + SR}) boosts performance further by encouraging the network to spot new boundary artifacts.

\begin{table}[t]
\centering  
\small
\begin{tabular}{p{1.5cm}ccccc} 
 Dataset & Carvalho & In-The-Wild & COVER  & CASIA \\ 
 \hline
CP + S &0.343 &0.430 & 0.351 & 0.242\\ 
CP + G + S &0.354 &0.441 &0.355 &0.270 \\
CP + GSR &\textbf{0.418} &\textbf{0.479} &\textbf{0.381} &\textbf{0.331} \\ 
\end{tabular}

\caption{\textbf{F1 score manipulation segmentation comparison trained with COCO annotations.}}
\label{tab:weakly}
\end{table}

\subsection{Qualitative Results}

\noindent\textbf{Generation Visualization}.
We illustrate some visualizations of the generation network in Figure~\ref{fig:gan}. It is clear that the generation network learns to match the pasted region with background during training. As a result, the boundary artifacts are becoming subtle and the generation network produces harder examples for the segmentation network.

\noindent\textbf{Segmentation Results}.
We present qualitative segmentation results on four datasets in Figure~\ref{fig:qr}. Unsurprisingly, the boundary branch outputs the potential boundary artifacts in manipulated images and the other branch fills in the interior based on the predicted manipulated boundaries. The examples indicate that our approach deals well with both splicing and copy-move manipulation based on the manipulation clues from the boundaries.

\section{Conclusion}

We propose a novel segmentation framework that firstly utilizes a generation network to enable generalization across variety of manipulations. Starting from copy-pasted examples, the generation network generates harder examples during training. We also design a boundary artifact guided segmentation and refinement network to focus on manipulation artifacts rather than semantic content. Furthermore, the segmentation and refinement stage share the same weights, allowing for much faster inference. Extensive experiments demonstrate the generalization ability and effectiveness of GSR-Net on four standard datasets and show state-of-the-art performance. The manipulation segmentation problem is still far from being solved due to the large variation of manipulations and post-processing methods. Including more manipulation techniques in the generation network could potentially boost the generalization ability of the existing model and is part of our future research.

\section*{Acknowledgement}
This work was supported by the DARPA MediFor program under cooperative agreement FA87501620191, ``Physical and Semantic Integrity Measures for Media Forensics''. 

{\small
\bibliographystyle{ieee}
\bibliography{main}
}

\end{document}